%% file: main_arxiv.tex
\documentclass[a4paper]{article}
\usepackage{authblk}
\usepackage{epsfig}
\usepackage{subcaption}
\usepackage{calc}
\usepackage{amssymb}
\usepackage{amstext}
\usepackage{amsmath}
\usepackage{amsthm}
\usepackage{multicol}
\usepackage{pslatex}

\usepackage{algorithm2e}
\usepackage[bottom]{footmisc}

\usepackage{amssymb, amsthm, amsfonts} 
\usepackage{color, bm, caption,  xcolor}
\usepackage{mleftright, xspace}

\newcommand{\E}[2]{\mathbb{E}_{#1}\mleft[#2\mright]}
\newcommand{\KL}[2]{KL\mleft(#1||#2\mright)}
\newcommand{\ph}{\psi}

\newcommand{\qh}[1]{q_{\ph}\mleft(#1\mright)}

\newcommand{\psr}[1]{p_{SR}\mleft(#1\mright)}

\newcommand{\z}{\bm{z}}
\newcommand{\zk}{\z_{<k}}
\newcommand{\x}{\bm{x}}
\newcommand{\xt}{\bm{\tilde{x}}}
\newcommand{\y}{\bm{y}}

\newcommand{\pt}[1]{p_\theta\mleft(#1\mright)}

\newcommand{\qp}[1]{q_\phi\mleft(#1\mright)}

\newcommand{\pd}[1]{p_{\mathcal{D}}\mleft(#1\mright)}
\newcommand{\vdvae}{VD-VAE\xspace}
\newcommand{\cvdvae}{CVDVAE\xspace}

\newcommand{\ny}{m}
\newcommand{\rmse}[2]{\frac{1}{\sqrt{\ny}}\| #1 - #2\|_2}
\newcommand{\Hs}{H_s}

\newcommand{\mutl}[1]{\mu_{\theta, l}\left(#1\right)}

\newcommand{\Vtil}[1]{
    \ifthenelse{\equal{#1}{}}{\Sigma_{\theta, l}^{-1}}{\Sigma_{\theta, l}^{-1}\left(#1\right)}
    }
\newcommand{\Vpil}[1]{
    \ifthenelse{\equal{#1}{}}{\Sigma_{\phi, l}^{-1}}{\Sigma_{\phi, l}^{-1}\left(#1\right)}
}
\newcommand{\mupl}[1]{
    \ifthenelse{\equal{#1}{}}{\mu_{\phi, l}}{\mu_{\phi, l}\left(#1\right)}
    }
\newcommand{\Vptil}[1]{
    \ifthenelse{\equal{#1}{}}{\Sigma_{\phi, \tau, l}^{-1}}{\Sigma_{\phi, \tau, l}^{-1}\left(#1\right)}
}
\newcommand{\Vptl}[1]{
    \ifthenelse{\equal{#1}{}}{\Sigma_{\phi, \tau, l}}{\Sigma_{\phi, \tau, l}\left(#1\right)}
}
\newcommand{\muptl}[1]{
    \ifthenelse{\equal{#1}{}}{\mu_{\phi, \tau, l}}{\mu_{\phi, \tau, l}\left(#1\right)}
    }

\newtheoremstyle{TheoremNum}
{\topsep}{\topsep}              
{\itshape}                      
{}                              
{\bfseries}                     
{.}                             
{ }                             
{\thmname{#1}\thmnote{ \bfseries #3}}
\theoremstyle{TheoremNum}

\usepackage{url}

\newcommand{\R}{\mathbb{R}}
\date{}
\title{Efficient Posterior Sampling For Diverse Super-Resolution with Hierarchical VAE Prior}

 \author[1]{Jean Prost}
 \author[2]{Antoine Houdard}
 \author[3]{Andrés Almansa}
 \author[4]{Nicolas Papadakis}
 \affil[1]{Univ. Bordeaux, Bordeaux IMB, INP, CNRS, UMR 5251, F-33400 Talence, France}
 \affil[2]{Ubisoft La Forge, F-33000 Bordeaux}
 \affil[3]{Université Paris Cité, CNRS, MAP5, F-75006 Paris}
 \affil[4]{Univ. Bordeaux, CNRS, INRIA, Bordeaux INP, IMB, UMR 5251, F-33400 Talence, France}
\begin{document}


\maketitle
\input{abstract.tex}

 \section{\uppercase{Introduction}}
 \label{sec:introduction}
 \input{introduction.tex}

\section{\uppercase{Hierarchical VAE}}
\label{sec:HVAE}
\input{background.tex}

\section{\uppercase{Super-Resolution with HVAE}}
\label{sec:super-resolution}
\input{super-resolution.tex}

\section{\uppercase{Experiments}}
\label{sec:experiments}
\input{experiments.tex}

\section{\uppercase{Related works}}
\label{sec:related_works}
\input{related_works.tex}

\section{\uppercase{Conclusions}}
\label{sec:conclusion}
\input{conclusions.tex}\vspace{-0.25cm}

\section*{ACKNOWLEDGEMENTS}
 This study has been carried out with financial support from the French Research Agency through the PostProdLEAP project (ANR-19-CE23-0027-01).
 Computer experiments for this work ran on several platforms including HPC resources from GENCI-IDRIS (Grant 2021-AD011011641R1), and the PlaFRIM experimental testbed, supported by Inria, CNRS (LABRI and IMB), Université de Bordeaux, Bordeaux INP and Conseil Régional d’Aquitaine (see https://www.plafrim.fr).

\appendix
\input{appendix.tex}
\newpage
\bibliographystyle{abbrv}

\end{document}

%% file: abstract.tex
\abstract{
    We investigate the problem of producing diverse solutions to an image super-resolution problem. From a probabilistic perspective, this can be done by sampling from the posterior distribution of an inverse problem, which requires  the definition of a prior distribution on the high-resolution images. 
    In this work, we propose to use a pretrained hierarchical variational autoencoder (HVAE) as a prior.  We train a lightweight stochastic encoder to encode low-resolution images in the latent space of a pretrained HVAE. At inference, we combine the low-resolution encoder and the pretrained generative model to super-resolve an image. We demonstrate on the task of face super-resolution that our method provides an advantageous trade-off between the computational efficiency of conditional normalizing flows techniques and the sample quality of diffusion based methods.
}

%% file: introduction.tex
\begin{figure*}
    \centering
    \includegraphics[width=.87\linewidth]{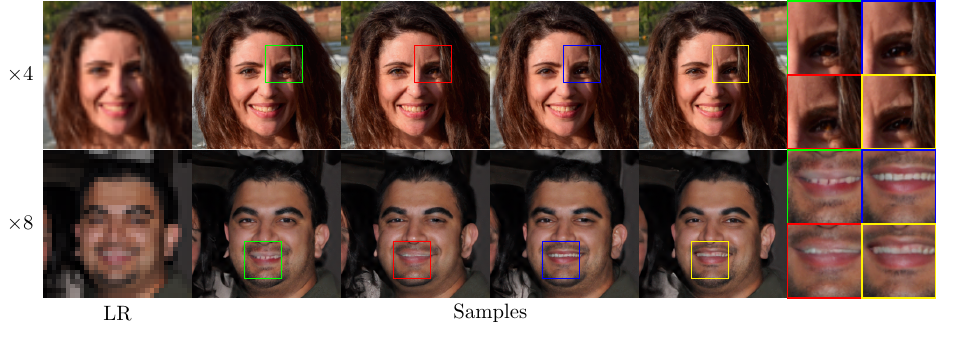}\vspace{-0.2cm}
    \caption{Diverse super-resolved samples at different upscaling factors.
    Our method can generate realistic samples with diverse attributes (hairs, mouth, eyes\dots).\vspace{-0.2cm}\label{fig:sr_samples}}
\end{figure*}

Image super-resolution is the task of generating a high-resolution (HR) image $\x$ corresponding to a low-resolution (LR) observation $\y$. A typical approach for image super-resolution is to train a deep neural network in a supervised fashion to map a LR image to its HR counterpart (see~\cite{lepcha2022image} for an extensive review). Despite impressive performances, those regression based methods are fundamentally limited by their lack of diversity. Indeed, there might exist many plausible HR solutions associated with one LR observation, but regression based methods only provide one of those solutions.

An alternative approach for image super-resolution is to sample from the posterior distribution $p(\x|\y)$. Specifically, we can 
train conditional deep generative models to fit the posterior $p(\x|\y)$. With the recent advances in deep generative modeling, it is possible to generate realistic and diverse samples from the posterior distribution. 

Starting from the seminal work of ~\cite{lugmayr2020srflow}, many approaches proposed to train conditional generative models such as conditional normalizing flow or conditional variational autoencoders in order to model the posterior distribution of the super-resolution problem. We refer to those methods as \textbf{direct methods}, as they only require one network function evaluation (NFE) to generate one sample. 

With the recent development of score-based generative models (also known as denoising diffusion models)~\cite{ho2020denoising,song2021scorebased}, posterior sampling methods based on conditional denoising diffusion models are now able to produce high-quality samples outperforming previous direct methods~\cite{choi2021ilvr,chung2022diffusion,kawar2022denoising}. 
However, denoising diffusion methods are limited by their computationally expensive sampling process, as they require numerous network function evaluations to generate one super-resolved sample. In the following, we classify those methods as {iterative methods}. 

In this work, we address the question: {Can we get the best of both worlds between the sampling quality of iterative methods, and the computationnal efficiency of direct methods?} We show that it is indeed possible to reach this goal with our diverse super-resolution method \cvdvae (Conditional VDVAE). 

Our approach is based on reusing a pretrained hierarchical variational autoencoder (HVAE)~\cite{LVAE,kingma2016improved}. HVAE models are able to generate high-quality images by relying on an expressive sequential generative model~\cite{vahdat2020nvae,child2021deep,hazami2022less}. By associating one latent variable subgroup to each residual block of a generative network, HVAE models are able to learn compact high-level representations of the data, and they can generate new samples efficiently, with only one evaluation of the generative network.

The fast sampling time and the expressivity of HVAE models make them suitable candidates for efficient posterior sampling. In this work, we exploit a pretrained \vdvae model~\cite{child2021deep}.  In order to repurpose \vdvae generative model for image super-resolution,
we train a low-resolution encoder to encode LR images in the latent space of the \vdvae model. 
By combining the LR encoder with the \vdvae generative model, we can  produce a sample with only one (autoencoder) network evaluation. 
By adopting a stochastic model for the LR encoder, our method can generate diverse samples from the posterior distribution, as illustrated in Figure~\ref{fig:sr_samples}. We show that the LR encoder can be trained with reasonable computational resources by exploiting the \vdvae original (HR) encoder to generate labels for training the LR-encoder, and by sharing weights between the LR encoder and \vdvae generative model.
We evaluate our method on super-resolution of face images, with upscaling factor $\times 4$ and $\times 8$ and  demonstrate that it reaches sample quality on par with sequential methods, while being significantly faster ($>\times 500$).

The  paper is organised as follows. In section~\ref{sec:HVAE}, we provide the necessary background on HVAE models. Then we present in section~\ref{sec:super-resolution} our super-resolution method. Experimental results are given in section~\ref{sec:experiments} and we discuss related works in section~\ref{sec:related_works}.

%% file: background.tex
{\bf Variational autoencoder}
We propose to use a hierarchical variational autoencoder as a prior model over high-resolution images. A variational autoencoder is a deep latent variable model of the form:
\begin{equation}
    \label{eq:marginal_x}
    \pt{\x} = \int \pt{\z}\pt{\x|\z} d\z.
\end{equation}
where $\pt{\z}$ defines the prior distribution of the latent variable $\z$ and  $\pt{\x|\z}$ is the decoding distribution.

A VAE also provides an inference model (encoder) $\qp{\z|\x}$, trained to match the intractable model posterior $\pt{\z|\x}$~\cite{kingma2013auto}. In order to define expressive models, both the generative model $\pt{\z, \x}$ and the encoder $\qp{\z|\x}$ are parameterized by neural networks, whose weights are respectively parameterized with $\theta$ and $\phi$.\newline

\begin{figure*}
    \begin{subfigure}[b]{0.55\linewidth}
        \centering
        \includegraphics[width=0.7\linewidth]{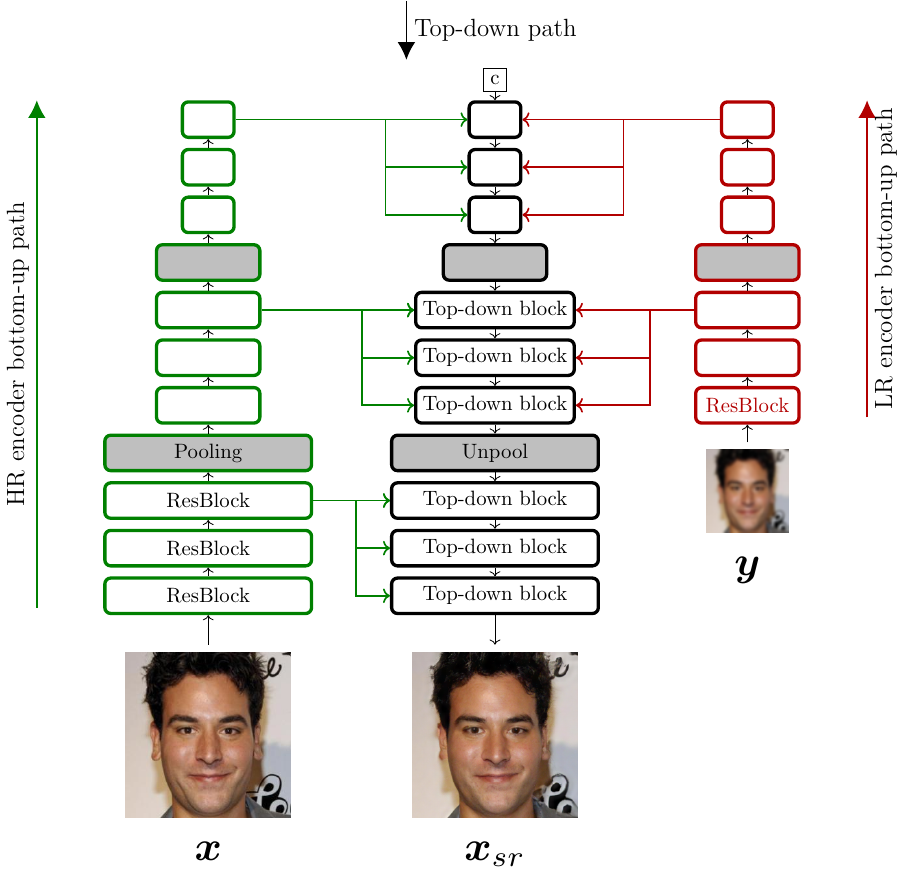}\vspace{-0.2cm}
        \caption{\cvdvae architecture \label{fig:fullnet}}
    \end{subfigure}
    \begin{subfigure}[b]{0.4\linewidth}
        \centering
        \includegraphics[width=.9\linewidth]{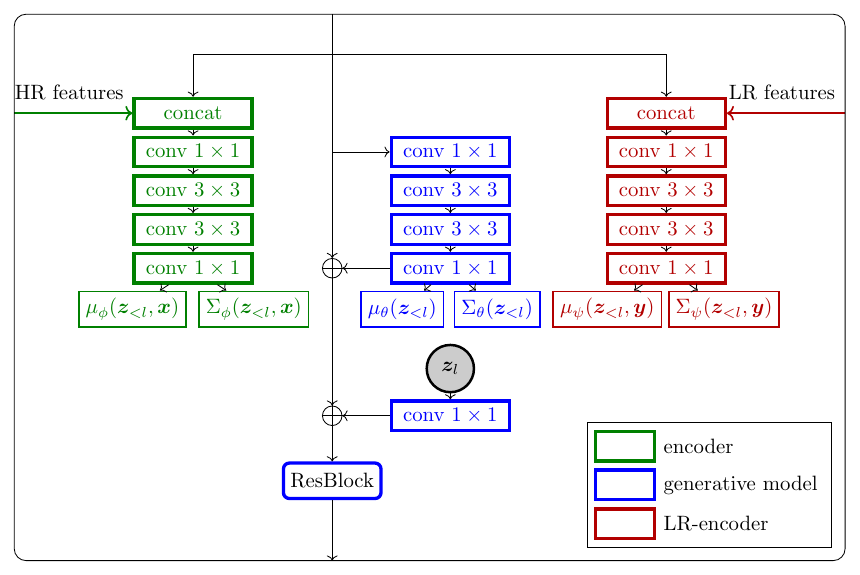}
        \caption{Top-down block\label{fig:tdblock}}
    \end{subfigure}
    \hspace{5pt}
    \caption{Super-resolution model based on a pretrained \vdvae model. (a) Given a pretrained \vdvae network, composed of an encoder bottom-up network (in green), and a top-down network (in black), we train a LR encoder (in red) to encode low-resolution images in the latent space of \vdvae. The LR-encoder has the same structure than \vdvae encoder, with a bottom-up path that extracts multiscale features, and a top-down path merged with \vdvae top-down path. (b) In each block of the top-down path, we add  branch for the LR encoder (in red) to infer the statistics of $\qh{\z_l|\z_{<l}, \y}$. \label{fig:vdvae_sr}
    }
\end{figure*}

\noindent
{\bf Hierarchical generative model}
A hierarchical VAE is a specific class of VAE where the latent variable $\z$ is partitioned into $L$ subgroups $\z = (\z_0, \z_1, \cdots, \z_{L-1})$, and the prior is set to have a hierarchical structure:
\begin{align}
    \pt{\z} & = \pt{\z_0, \z_1, \cdots, \z_{L-1}} \\
    &= \pt{\z_0} \prod_{l=1}^{L-1} \pt{\z_l|\z_{<l}}.\label{eq:hierarchical_prior}
\end{align}
In practice, each latent supgroup is a 3-dimensional tensor $\z_l \in \R^{c_l \times h_l \times w_l}$, with increasing resolution $h_0 \leq h_1 \leq \cdot \leq h_{L-1}$. Each conditional model in the hierarchical prior is set as a Gaussian: 
\begin{equation}
    \label{eq:Gaussian_hvae_prior}
    \begin{cases}
        \pt{\z_0} &= \mathcal{N}\left(\z_0; \mu_{\theta, 0}, \Sigma_{\theta, 0}\right) \\
        \pt{\z_l|\z_{<l}} &= \mathcal{N}\left(\z_l; \mutl{\z_{<l}}, \Sigma_{\theta, l}(\z_{<l})\right).
    \end{cases}
\end{equation} 
As illustrated in Figure~\ref{fig:vdvae_sr}, the generative model is embedded within a "top-down" generative network. 
To generate an image, a low-resolution constant input tensor is sequentially processed by a serie of top-down blocks and upsampling layers (Figure~\ref{fig:fullnet}). In each top-down block $l$ (Figure~\ref{fig:tdblock}), a latent subgroup $\z_l$ is sampled according to the statistics $\mutl{\z_{<l}}$ and $\Sigma_{\theta, l}(\z_{<l})$ computed within the top-down block.\newline

\noindent
{\bf Hierarchical encoder}
The HVAE encoder has the same hierarchical structure as the generative model:
\begin{equation}
	\label{eq:henc}
	\qp{\z|\x} = \qp{\z_0|\x}\prod_{l=1}^L \qp{\z_l|\z_{<l}, \x},
\end{equation}
with Gaussian parameterization of the conditional distributions:
\begin{equation}
    \label{eq:gaussian_hvae_encoder0}
    \hspace*{-0.3cm}\begin{cases}
         \qp{\z_0|\x}\hspace{-0.3cm}&= \mathcal{N}\left(\z_0; \mu_{\phi, 0}(\x), \Sigma_{\phi, 0}(\x)\right) \\
         \qp{\z_l|\z_{<l},\x} \hspace{-0.3cm}&= \mathcal{N}\left(\z_l; \mupl{\z_{<l}, \x}, \Sigma_{\phi, l}(\z_{<l}, \x)\right)\hspace{-0.1cm}. \\        
     \end{cases}
\end{equation}
The HVAE encoder is merged with the generative top-down network. In each top-down block $l$, a branch associated with the encoder uses features from the input $\x$ along with features from the previous levels to infer the encoder statistics at level $l$ (in green in Figure~\ref{fig:tdblock}). The image features are extracted by a "bottom-up" network (in green in Figure~\ref{fig:fullnet}).

%% file: super-resolution.tex
{\bf Problem formulation}
In this section we describe our super-resolution method based on a pretrained HVAE model.
We assume that the LR image $\y \in \mathbb{R}^{3\times \frac{H}{s}\times\frac{W}{s}}$, and its associated HR image $\x \in \mathbb{R}^{3\times H \times W}$ are related by a linear degradation model:
\begin{equation}
    \label{eq:forward}
    \y = (k * \x)\downarrow_s,
\end{equation}
where $k$ is a low-pass filter and $(u)\downarrow_s$ is defined as the subsampling operation with downsampling factor $s$.
Our goal is to sample from the posterior distribution of the inverse problem:
\begin{equation}
    \label{eq:posterior}
    p(\x|\y) \propto p(\y|\x)p(\x).
\end{equation}
In~\eqref{eq:posterior}, the likelihood $p(\y|\x)$ can be deduced from the degradation model~\eqref{eq:forward}. On the other hand, the prior model $p(\x)$ needs to be specified by the user. Deep generative models such as GANs, VAE or diffusion models can be used to model the prior on high-resolution images. In the following, we propose to parameterize $p(\x)$ with a hierarchical variational autoencoder. Given a pretrained HVAE prior $\pt{\x}$, the ideal super-resolution model is:
\begin{equation}
    \label{eq:posterior2a}
    \pt{\x|\y} = \int \pt{\x|\y, \z}\pt{\z|\y} d \z,
\end{equation}
where probability laws correspond to the conditional of the augmented model $\pt{\z, \x, \y} := \pt{\z}\pt{\x|\z}p(\y|\x)$. 
Since we do not have acces to $\pt{\x|\y, \z}$ and $\pt{\z|\y}$, we can not directly sample from~\eqref{eq:posterior2a}. However, we will see in the following part that we can efficiently approximate this model by making use of the structure of the HVAE hierarchical latent representation and of the its pretrained encoder.\newline

\noindent
{\bf Super-resolution model}
It has been observed in several works that the low-frequency information of images generated by HVAE model where mostly controlled by the low-resolution latent variable, at the beginning of the hierarchy~\cite{vahdat2020nvae,child2021deep,Havtorn2021HierarchicalVK}. Preliminary experiments, detailed in  appendix~\ref{app:details_uk} validate those previous observations.
Hence, for a large enough number of latent groups $k$, samples from $\pt{\x|\x_{<k}}$ share the same low-frequency information. As a consequence, all the samples from $\pt{\x|\x_{<k}}$ are consistent to a LR image $\y$ (up to a small error). 
This motivates us to define the following super-resolution model:
\begin{equation}
    \label{eq:psr}
    \psr{\x|\y} = \int \pt{\x|\z_{<k}}\qh{\z_{<k}|\y} d \z,
\end{equation}
where $\qh{\z|\y}$ is a stochastic low-resolution encoder, trained to encode the low-resolution latent groups. By definition of the super-resolution model~\eqref{eq:psr}, we can sample from $\psr{\x|\y}$ by sequentially sampling $\z_{<k} \sim \qh{\z_{<k}|\y}$ and $\x \sim \pt{\x|\z_{<k}}$.\newline

\noindent
{\bf Hierarchical low-resolution encoder}
We set the LR encoder to have a hierarchical structure:
\begin{equation}
    \label{eq:lr_enc}
    \qh{\z_{<k}|\y} = \qh{\z_0|\y}\prod_{l=1}^{k} \qh{\z_l|\z_{<l}, \y},
\end{equation}
with Gaussian conditional distributions:
\begin{equation}
    \label{eq:gaussian_hvae_encoder}
    \begin{cases}
         \qh{\z_0|\y} &= \mathcal{N}\left(\z_0; \mu_{\psi, 0}(\y), \Sigma_{\psi, 0}(\y)\right) \\
         \qh{\z_l|\z_{<l},\y} &= \mathcal{N}\left(\z_l; \mu_{\psi,l}(\z_{<l}, \y), \Sigma_{\psi, l}(\z_{<l}, \y)\right). \\        
     \end{cases}
\end{equation}
We implement the LR encoder with the same architecture as \vdvae original (HR) encoder, but with a limited number of blocks due to reduced number of latent variable to be predicted (Figure~\ref{fig:vdvae_sr}). 
Only the parameters of the low-resolution encoder (in red in Figure \ref{fig:vdvae_sr}) are trained, while the shared parameters (in blue in Figure \ref{fig:vdvae_sr}) are set to the value of the corresponding parameters in the pretrained \vdvae generative model, and remain frozen during training.\newline

\noindent
{\bf Training}
We keep the weights of the HVAE decoder $\pt{\x|\z}$, so that the only trainable weights of our super-resolution model~\eqref{eq:psr} are the weights of the LR encoder $\psi$. Given a joint training distribution of HR-LR image pairs $\pd{\x,\y}$, the LR encoder is trained to match the available "high-resolution" HVAE encoder $\qp{\z_{<k}|\x}$ on the associated HR images, by minimizing the Kullback-Leibler (KL) divergence:
\begin{equation}
    \label{eq:loss}
    \mathcal{L}(\ph) = \E{\pd{\x, \y}}{\KL{\qp{\z_{<k}|\x}}{\qh{\z_{<k}|\y}}}.
\end{equation}
The criterion~\eqref{eq:loss} was introduced by~\cite{harvey2022conditional}, who demonstrated that minimizing~\eqref{eq:loss}  is equivalent to maximizing a lower-bound of the super-resolution conditional log-likelihood on the training dataset, and that under additional assumptions on the pretrained HVAE model, one can  reach optimal performance by only training the low-resolution encoder $\qh{\z_{<k}|\y}$.
In practice, the KL divergence within the training criterion~\eqref{eq:loss} can be decomposed into a sum of KL divergence on each latent subgroup:
\begin{equation}
    \label{eq:kl_decomp}
    \begin{split}
            &\KL{\qp{\z_{<k}|\x}}{\qh{\z_{<k}|\y}} = \KL{\qp{\z_0|\x}}{\qh{\z_0|\y}} \\ &+ \E{\qp{\z_{<k}|\x}}{
        \sum_{l=1}^{k} \KL{\qp{\z_l|\z_{<l}, \x}}{\qh{\z_l|\z_{<l},\y}}
    }.	
    \end{split}
\end{equation}
Since each conditional law involved in~\eqref{eq:kl_decomp} is Gaussian, each KL term can be computed in closed-form. In practice the covariance matrices $\Sigma_{\phi, l}(\z_{<l}, \x)$ and $\Sigma_{\psi, l}(\z_{<l}, \y)$ are constrained to be diagonal, so that the KL can be computed efficiently.

%% file: experiments.tex
\subsection{Experimental settings}
\paragraph*{Dataset and upscaling factors}
We test our super-resolution method \cvdvae on the FFHQ dataset~\cite{karras2019style}, with images of resolution $256\times256$.  
We experiment on 2 upscaling factors: $\times 4$ ($64\times64 \rightarrow 256 \times 256$) and $\times 8$ ($32\times32 \rightarrow 256 \times 256$).
The low resolution images are initially downscaled by applying an antialiasing kernel followed by a bicubic interpolation.
\newline

\noindent
{\bf Compared methods}
We compare \cvdvae with a conditional normalizing flow (HCFlow)~\cite{liang2021hierarchical}, a conditional diffusion model (SR3)~\cite{saharia2021image}, and a method that add guidance to a non-conditional diffusion model at inference (DPS)~\cite{chung2022diffusion}.
We retrain HCFlow on FFHQ256 using the official implementation.
For DPS, we also reuse the official implementation with the available pretrained model, which was trained on FFHQ.
For SR3, since no official implementation is available, we used an open-source (non-official) implementation~\cite{janspiry2022sr3}, and we trained a model on FFHQ. When training SR3, we found  that a color shift~\cite{deck2023easing} was responsible for important reconstruction errors. To compensate this weakness of the method, we project the super-resolved image on the space of consistent solutions at inference as proposed in~\cite{bahat2020explorable}.
For fair comparison, we retrained both HCFlow and SR3 with the same computational budget as for our low-resolution encoder. 
For HCFlow and \cvdvae, we set the temperature of the latent variables at $\tau=0.8$ during sampling.
\newline

\noindent
{\bf
Evaluating a diverse SR method}
Due to the ill-posedness of the problem, evaluating a diverse super-resolution model based solely on the distortion to the ground truth is not satisfactory.
 Indeed, there exist many solutions that are both realistic and consistent with the LR input while being far from the ground truth. 
 Thus, in order to evaluate the super-resolution model, we provide a series of metrics that evaluate different expected characteristics of a diverse super-resolution model, 
such as the consistency of the solution, the diversity of the samples and the general visual quality.   
It should be noted that those metrics are not necessarily correlated: a model could generate diverse solutions, that are not consistent or realistic, or, on the opposite, it could provide solutions that are realistic and consistent but with a low diversity.
Thus, to evaluate a diverse super-resolution model, it is necessary to consider these three different aspects  together: diversity, consistency and visual quality. \newline

\noindent
{\bf Evaluation metrics}
The general quality of the super-resolved images is evaluated using the blind Image quality metric BRISQUE~\cite{mittal2012no}. 
Consistency with the LR input is measured via peak Signal-to-Noise Ratio (PSNR, denoted as LR-PSNR in Tables~\ref{tab:metrics} and~\ref{table:effect_temp}).
Furthermore, to evaluate the diversity of the super-resolution, we compute the Average Pairwise distance  between different samples coming from the same LR input (denoted as APD in Tables~\ref{tab:metrics} and~\ref{table:effect_temp}),
both at the pixel level, using the mean square error (MSE) between samples (considering pixel intensity value between 0 and 1), and at a perceptual level using  the  LPIPS similarity criteria~\cite{zhang2018perceptual}.
For one LR input, the average pairwise distance is computed as the average distance between all the possible pairs of images in a set of 5 super-resolved samples. The reported APD  in Tables~\ref{tab:metrics} and~\ref{table:effect_temp} corresponds to the mean value of the single image APD over 500 LR inputs in the test set.
We measure the distortion of the super-resolved samples with respect to the ground truth HR image in terms of PSNR,
structural similarity (SSIM)~\cite{wang2004image} and LPIPS, as it is common in the super-resolution literature. 
All numbers reported correspond to the metric mean value on a subset of 1000 images from FFHQ256 test set.

\subsection{Results}

\paragraph{Quantitative evaluation}

\begin{table*}[!t]

    \centering
    \includegraphics[width=.9\linewidth]{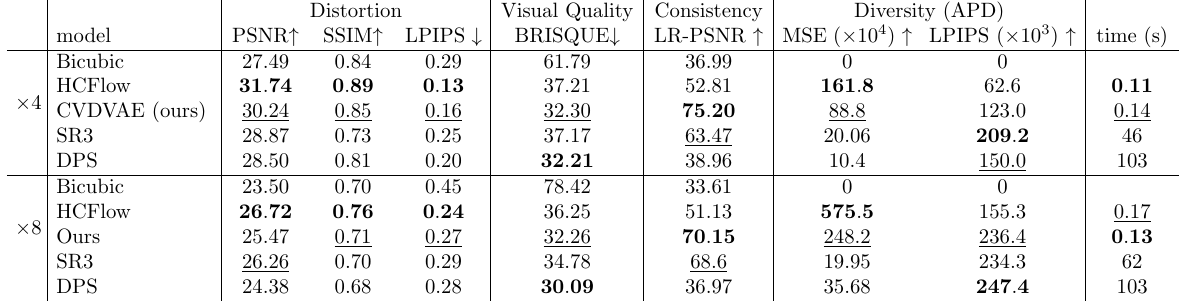}
        \caption{\label{tab:metrics}Comparison of diverse SR methods face super-resolution. 
    Best result is in bold, second best is underlined.}
\end{table*}
The quantitative results presented in Table~\ref{tab:metrics} indicate that \cvdvae 
provides a good trade-off between the different evaluated metrics. Indeed, it obtains the second best results in terms of distortion and visual quality, and the second or third best results in terms of diversity. \cvdvae is also one of the fastest methods, along with HCFlow.
HCFlow provides the best results for distortion metrics as it explicitly penalizes bad reconstruction in its training loss. Similar to CVDVAE, its application is fast, as it requires only one network evaluation to produce a super-resolved image. However, HCFlow lacks high-level diversity (as measured by the LPIPS average pairwise distance), compared with the concurrent methods. We postulate that this lack of diversity is due to the relative lack of expressiveness of normalizing flows architecture compared to the convolutional architectures used by diffusion and HVAE models.
Our method, along with DPS, produces the best results in terms of visual quality as measured by the BRISQUE metric, illustrating the benefit of using a pretrained unconditional generative model. The computational cost of DPS is nevertheless significantly higher than the ones of  CVDVAE and HCFlow, as DPS requires $1000$ steps of network evaluations (and backpropagation through the denoiser) to produce one super-resolved sample. Finally, SR3 performances are inferior to the compared methods. We used the same computational budget (48 hours on 4 GPUs) for training the SR3 models than our CVDVAE and HCFlow. This computational budget is significantly lower than the one reported in the SR3 paper~\cite{saharia2021image}
($\approx 4$ days on 64 TPUv3 chip), 
 and we expect that training the SR3 model for more epochs would improve its performance. Like DPS, SR3 is slower than our method as it requires 2000 network evaluations to produce one super-resolved image; although, unlike DPS, SR3 does not require to backpropagate through the score-network.
\newline

\begin{figure}
    \begin{subfigure}[h]{\linewidth}
        \centering
        \includegraphics[width=.8\linewidth]{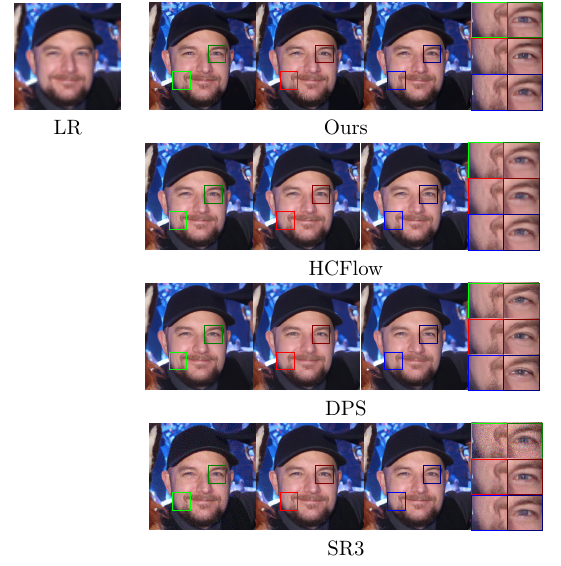}
    \end{subfigure}
    \begin{subfigure}[h]{\linewidth}
        \centering
        \includegraphics[width=.8\linewidth]{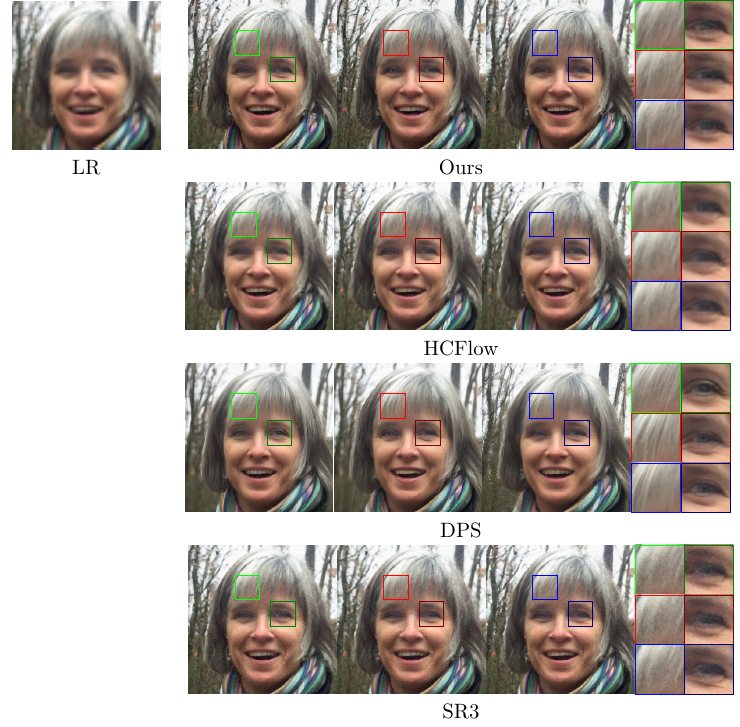}
    \end{subfigure}    
    \caption{Samples from different diverse SR methods ($\times 4$)\label{fig:comp_diverse_x4}}
\end{figure}

\begin{figure}
    \begin{subfigure}[h]{\linewidth}
        \centering
        \includegraphics[width=.8\linewidth]{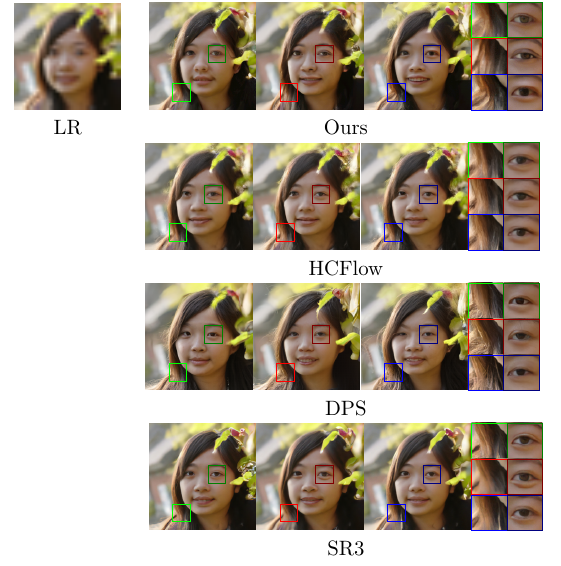}
    \end{subfigure}
    \begin{subfigure}[h]{\linewidth}
        \centering
        \includegraphics[width=.8\linewidth]{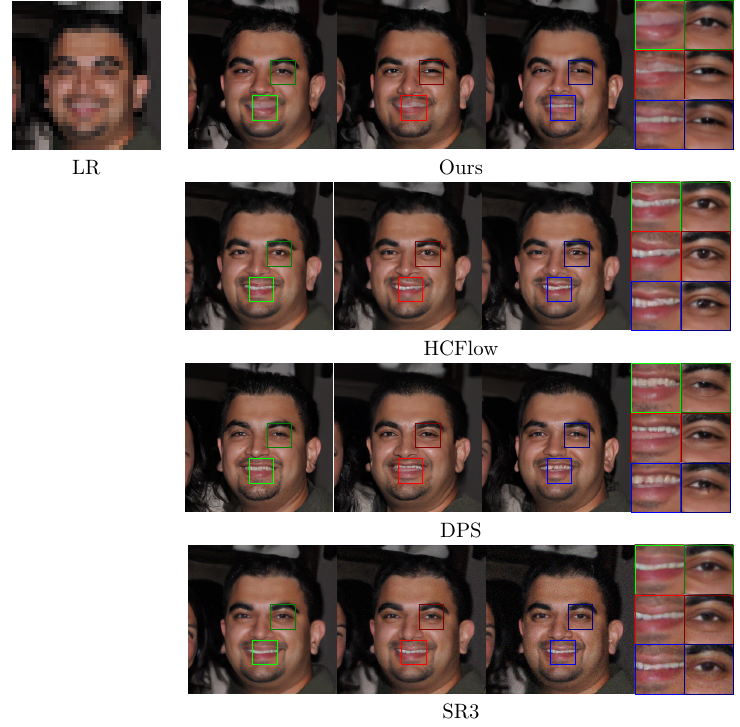}
    \end{subfigure}    
    \caption{Samples from different diverse SR methods ($\times 8$)\label{fig:comp_diverse_x8}}
\end{figure}

\noindent
{\bf Qualitative evaluation}
Visual comparisons of super-resolved samples from the different evaluated methods are provided in Figures~\ref{fig:comp_diverse_x4} and~\ref{fig:comp_diverse_x8}. 
CVDVAE is able to produce diverse textures as illustrated by the facial hair variation in Figure~\ref{fig:comp_diverse_x4} or the hair variation in~\ref{fig:comp_diverse_x8}.
\cvdvae appears to produce super-resolved samples with higher semantic diversity, in terms of textures (hairs, skin), in line with the higher perceptual diversity measured in the quantitative evaluation.\newline

\begin{figure*}
    \centering
    \includegraphics[scale=1.15]{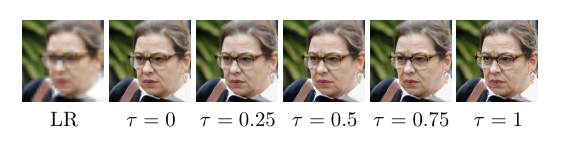}\vspace{-0.45cm}
    \caption{Effect of the sampling temperature $\tau$ on the super-resolved result. Increasing the temperature yields image with more high-frequency details.\vspace{-0.1cm}}
    \label{fig:demo_temp}
\end{figure*}

\noindent
{\bf Temperature control}
As for the unconditionnal HVAE models, \cvdvae offers the possibility to control the conditional generation via the temperature of the latent variable distributions~\cite{vahdat2020nvae,child2021deep}. The temperature parameter $\tau$ controls the variance of the Gaussian latent distributions. 
In order to assess the behavior of the model on both low and high temperature regime, we evaluate our method on 2 temperatures ($\tau \in \{0.1, 0.8\}$). 
Quantitative results in Table~\ref{table:effect_temp} show that reducing the temperature leads to a solution closer to the ground truth in terms of low-levels distortion metrics (PSNR and the SSIM), while using a higher temperature helps to improve the perceptual similarity (LPIPS) with the ground-truth, as well as the general perceptual quality of the generated HR images and the diversity of the samples.
On Figure \ref{fig:demo_temp}, we display \cvdvae's samples at different temperatures $\tau$.
The sampling temperature correlates with the perceptual smoothness of the super-resolved sample, a higher sampling temperature inducing images with sharper details.

 \begin{table*}
    \centering
  
    \includegraphics*[width=.82\linewidth]{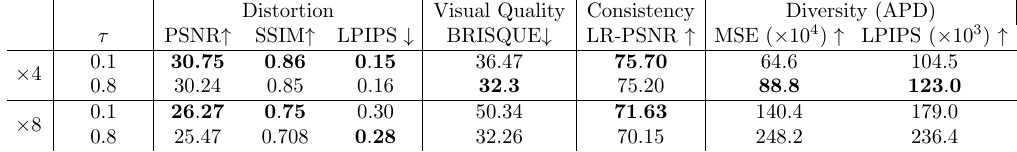}\vspace{-0.1cm}
      \caption{Effect of the sampling temperature $\tau$ on CVDVAE super-resolution results.\vspace{-0.4cm}\label{table:effect_temp}}
 \end{table*}   

%% file: related_works.tex
\paragraph*{Super-resolution with pre-trained generative models}
A large number of methods were designed to solve imaging inverse problems such as image super-resolution by using pretrained deep generative models (DGM) as a prior.
 This includes methods relying on generative adversarial networks (GAN)~\cite{menon2020pulse,marinescu2020bayesian,pan2021exploiting,daras2021intermediate,daras2022score,poirier2023robust}, variational autoencoders~\cite{mattei2018leveraging,gonzalez2021solving,Prost_2023_ICCV} and denoising diffusion models~\cite{choi2021ilvr,chung2022diffusion,kawar2022denoising,Song2023-pigdm}.
However, those approaches are computationally expensive as they require an iterative sampling or optimization procedure which require many network evaluation.
On the other hand, our approach enables fast inference (one network evaluation), at the cost of reduced flexibility (due to the need of training a task-specific encoder). 
The idea of training an encoder to map a degraded image in the latent space of a generative network was previously exploited in the context of image inpainting with HVAE~\cite{harvey2022conditional}, and super-resolution with a GAN prior~\cite{chan2021glean,richardson2021encoding}. 

\paragraph*{Diverse super-resolution with conditional generative models}
Although it is possible to sample from the posterior $p(\x|\y)$ by using an unconditionnal deep generative models, those methods are restricted to specific dataset for which pretrained models are available. On generic natural images, the state of the art methods rely on conditional generative models, directly trained to model the posterior $p(\x|\y)$~\cite{lugmayr2021ntire,lugmayr2022ntire}. Those methods include conditional normalizing flows~\cite{lugmayr2020srflow,liang2021hierarchical}, conditional GAN~\cite{bahat2020explorable}, conditional VAE~\cite{gatopoulos2020super,zhou2021vspsr,chira2022image} and conditional denoising diffusion models~\cite{li2022srdiff,saharia2021image}.

%% file: conclusions.tex
In this work we presented \cvdvae, a method that realizes an efficient sampling from the posterior of a super-resolution problem, by combining a low-resolution image encoder with a pretrained \vdvae generative model. 
\cvdvae showed promising results on face super-resolution, on par with state-of-the-art diverse SR methods, providing semantically diverse and high-quality samples. Our results illustrate the ability of conditional hierarchical generative models to perform complex image-to-image tasks. 

Our results are in line with many works that illustrates the benefits of using HVAE models for downstream applications~\cite{Havtorn2021HierarchicalVK,agarwal2023a,Prost_2023_ICCV}.  One drawback of our approach is its limitation to dataset for which pretrained HVAE models are available, such as human faces or low-resolution ImageNet. 
 However, we postulate that HVAE models have not yet reached their limits, and, by adapting design features from current SOTA deep generative models~\cite{rombach2022high,kang2023scaling} (architectural improvement, longer training, larger dataset), HVAE models could significantly improve their performance and expreessiveness, and generalize on much diverse datasets. 

%% file: appendix.tex
\section{Low-frequency information in \vdvae latent represention}

\paragraph*{Average low-resolution pairwise distance between samples}
We propose to evaluate which latent variables encode  the low-frequency information within \vdvae generative model. We measure how close the image generated by the conditional models $\pt{\x|\zk}$ are with each other, when they are downsampled with different downscaling factors. Without loss of generality, we consider the root mean square error (RMSE) as a measure of distance between samples.
  Thus, we estimate:  
\begin{equation}
    \label{eq:uk}
  U_k^s := \mathbb{E}_{\pt{\z_{<k}}} \mathbb{E}_{\pt{\x|\z_{<k}}} \mathbb{E}_{\pt{\xt|\z_{<k}}} \left[ \rmse{\Hs\x}{\Hs\xt} \right],
\end{equation}
the average low-resolution pairwise distance of the generative model $\pt{\x|\zk}$, when samples are downsampled by a factor $s$. 
 $U_k^s$ measures to what extent images sampled from $\pt{\x|\zk}$ differ from each other when they are downsampled.\newline

\noindent
{\bf Estimating $U_k^s$} \label{app:details_uk}
We compute an estimations of the average sample low-resolution pairwise distance $U_k^s$ \eqref{eq:uk} with ancestral Monte-Carlo sampling. 
We sample $50$ different full latent codes $\z^{(i)}$ from the prior:
\begin{equation}
    \z^{(i)} \sim \pt{\z}.
\end{equation}
For each latent code $\z^{(i)}$ and each number of fixed groups $k$, $5$ we sample five images: 
\begin{equation}
\x^{(i,k,l)} \sim \pt{\x|\zk^{(i)}}.
\end{equation}
The average sample pairwise distance estimation is then computed as:
\begin{equation}
    \hat{U_k^s} = \sum_{i=1}^{50} \sum_{1\leq l < m \leq 5}\rmse{\Hs\x^{(i,k,l)}}{\Hs\x^{(i,k,m)}},
\end{equation}
where $\Hs$ is the downsampling operator associated to the downscaling factor $s$.\newline
\begin{figure}
    \centering
    \includegraphics[width=.5\linewidth]{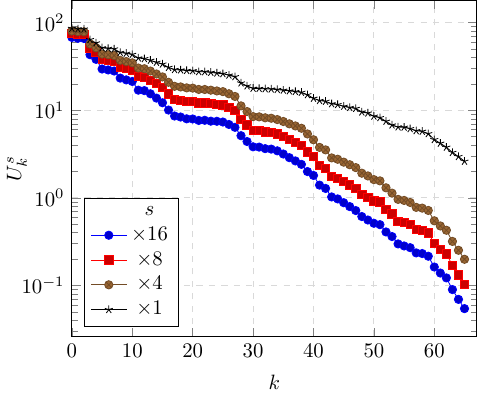}
    \caption{Average low-resolution pairwise distance, $U_k^s$ \eqref{eq:uk} between samples from the conditional generative model $\pt{\x|\z_{<k}}$ of \vdvae, for downscaling factors $s= 1, 4 , 8 , 16 $. Image with pixel values in $[0, 255]$.}
    \label{fig:uk}
\end{figure}

\noindent
{\bf Low-resolution consistency of \vdvae samples}

In Figure \ref{fig:uk} we estimate the value of $U_k^s$ for different downsampling factors. Results illustrate that, as the number of fixed groups $k$ increases, the generated images get more similar.
Furthermore, for a given number of fixed groups $k$, the low-resolution pairwise distance decreases as the downsampling factor $s$ increases, indicating that there is more variation in the HR samples than in their LR counterparts. 
The gap between the average sample pairwise distance in high resolution ($s=1$), and low-resolution ($s \in \{4, 8, 16\}$) gets larger as the number of fixed groups $k$ increases, indicating that fixing a large number of groups $k$ yields samples that are close at low-resolution but different at high-resolution.
The average low-resolution pairwise distance $U_k^s$ gets closer to zero as $k$ increases. While there is no value of $k$ such that $U_k^s=0$, we argue that for a large enough value of $k$, $U_k^s$ becomes negligible compared to the pixel intensity range (0-255), for instance $U_{60}^4 < 0.5$.
Those results show that the downsampling of any image synthesized from the conditional generative model $\pt{\x|\z_{<k}}$ are consistent with one low-resolution image $\y$ with a certain precision inversely proportional to $k$.